\documentclass{article}

\usepackage[preprint]{corl_2026} % Uncomment for pre-prints (e.g., arxiv); This is like ``final'', but will remove the CORL footnote.
\usepackage{times}
\usepackage{graphicx}
\usepackage{algorithm2e}
\usepackage{subcaption}
\usepackage{amsmath, amsfonts, amssymb}
\usepackage{tabularx}
\usepackage{booktabs}
\usepackage{multirow}
\usepackage{mathrsfs}
\usepackage[font=small]{caption}
\usepackage{threeparttable}
\usepackage[normalem]{ulem}
\usepackage{wrapfig}

\newcommand{\name}{SPOT\xspace}
\newcommand{\namelong}{\textbf{S}patial \textbf{P}erception-\textbf{O}riented Long-Horizon Humanoid \textbf{T}eleoperation}

\title{SPOT: \underline{S}patial \underline{P}erception-\underline{O}riented\\Long-Horizon Humanoid \underline{T}eleoperation}

\author{
  Lixing Fang$^{\ast,1}$, Ziyan Xiong$^{\ast,1}$, Sunli Chen$^1$, Zhiyang Dou$^2$, Chuang Gan$^{\dagger,1}$\\
  $^1$University of Massachusetts Amherst,\;$^2$Massachusetts Institute of Technology \\
}

\begin{document}
\maketitle

{
\let\thefootnote\relax
\footnotetext{$^\ast$equal contribution,\;$^\dagger$corresponding author}
}

\begin{center}
    \centering
    \captionsetup{type=figure}
    \vspace{-30pt}
    \includegraphics[width=0.94\textwidth]{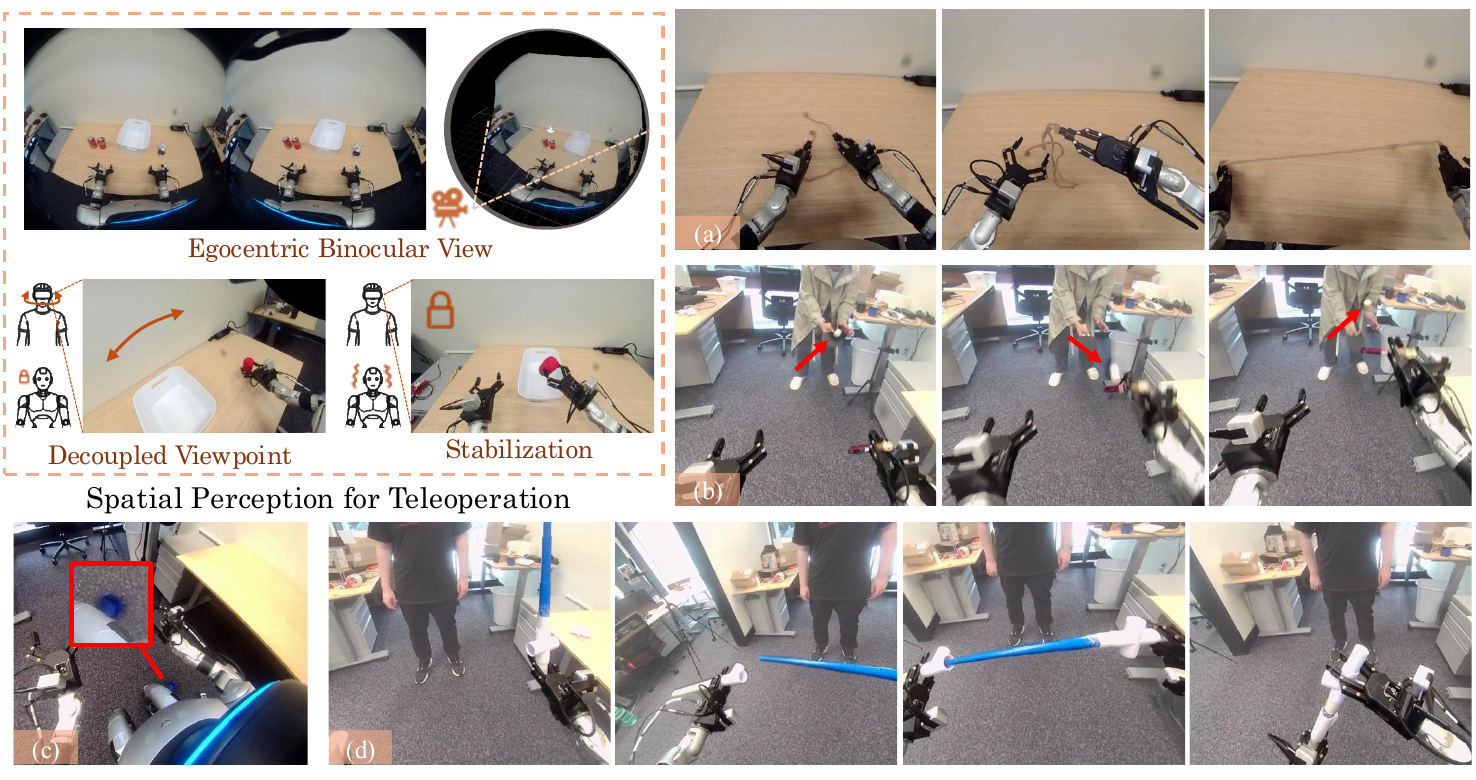}
    \vspace{-1mm}
    \caption{\name enhances spatial perception through: binocular egocentric visualization, viewpoint decoupling, and stabilization. It enables tasks including: (a) bimanual knot tying; (b) highly dynamic object interception; (c) dropped-object relocation; and (d) large-workspace bimanual manipulation with flexibility and ease of use.}
    \vspace{-5pt}
\label{fig:teaser}
\end{center}

%===============================================================================

\begin{abstract}
High-quality demonstration data is becoming a central bottleneck for training general-purpose humanoid robots. 
While recent humanoid teleoperation systems have made substantial progress in retargeting human motion to robot motion, long-horizon loco-manipulation requires another capability: operators must maintain task-relevant spatial awareness over time, e.g., object locations, surrounding environments, the robot's pose. We call the extent of this awareness the operator's \emph{perceptual horizon}. 
However, existing methods often shorten this: narrow views miss peripheral events, robot-mounted cameras become unstable during locomotion, and coupled head-view control makes looking around interfere with robot motion. 
We present \textbf{\name}, a Spatial Perception-Oriented VR Teleoperation system for collecting long-horizon humanoid demonstration data by providing extended perceptual horizon. \name combines a robot-mounted binocular fisheye camera, a wide-field stereoscopic display, viewpoint-decoupled free-looking, and visual stabilization to provide a robot-centric view that is wide, stable, and actively inspectable. 
Unlike conventional egocentric interfaces, \name decouples visual exploration from robot actuation: the egocentric stereo observation is rendered on a virtual hemisphere around the operator, so natural head rotations change where the operator looks within the wide-field view rather than commanding the robot head, camera, or torso. We evaluate \name on perception-critical humanoid data-collection tasks spanning drop recovery, peripheral retrieval, large-workspace bimanual manipulation, fine alignment, and dynamic interaction. 
\name improves efficiency, accuracy, and recovery speed, demonstrating its effectiveness for user-friendly and scalable long-horizon humanoid data collection.

% making teleoperation an increasingly important tool for large-scale data collection.

% We call the extent over which this awareness is reliably maintained the operator's perceptual horizon. 

% We present \name, a perception-centered VR teleoperation system that combines a robot-mounted binocular fisheye camera, a wide-field stereoscopic display, viewpoint-decoupled free-looking, and visual stabilization to provide a robot-centric view that is wide, stable, and actively inspectable. 
% Unlike prior systems, \name~treats spatial perception as an explicit interface design objective. \name~decouples visual exploration from robot actuation: the egocentric stereo observation is rendered on a virtual hemisphere around the operator, so natural head rotations change where the operator looks within the wide-field view rather than commanding the robot head, camera, or torso. We evaluate \name~on perception-critical humanoid tasks spanning drop recovery, peripheral retrieval, wide-range bimanual manipulation, fine alignment, and dynamic interaction. Results show improved efficiency, accuracy, recovery time, and reduced workload, highlighting perception transfer as key to scalable long-horizon humanoid teleoperation.

% suggesting that perception transfer is a key complement to motion transfer for scalable long-horizon humanoid data collection.

\end{abstract}


%===============================================================================

\section{Introduction}
\label{sec:introduction}
% Humanoid robots are rapidly entering a data-driven era. Recent advances in large-scale motion tracking, vision-language-action models, and teleoperation-based data collection have demonstrated the importance of high-quality demonstrations for acquiring whole-body humanoid skills. As a result, scalable collection of long-horizon loco-manipulation data is becoming an increasingly critical component of future humanoid learning systems. Existing teleoperation systems have achieved remarkable progress in motion transfer through accurate motion tracking, retargeting, and whole-body control, enabling dexterous manipulation and agile locomotion. However, successful long-horizon teleoperation requires more than faithfully reproducing human motions. Operators must continuously perceive and understand the surrounding environment, maintain accurate spatial understanding of task-relevant objects and events, coordinate locomotion and manipulation, and interact effectively with complex and evolving scenes throughout extended tasks.

High-quality demonstration data has become critical for training humanoid robots. As humanoids move from short manipulation episodes toward long-horizon loco-manipulation, teleoperation is increasingly used to collect data involving mobility, bimanual interaction, recovery from unexpected events, and precise spatial alignment~\cite{ze2025twist2scalableportableholistic,cheng2024open,he2024omnih2ouniversaldexteroushumantohumanoid,luo2026sonicsupersizingmotiontracking,li2025amoadaptivemotionoptimization}. Existing humanoid teleoperation systems have made substantial progress in transferring human motion to robot motion through motion tracking~\cite{he2024omnih2ouniversaldexteroushumantohumanoid,ben2025homiehumanoidlocomanipulationisomorphic,li2025cloneclosedloopwholebodyhumanoid,myers2025child,purushottam2025heavy}, retargeting~\cite{he2024learning,ze2025twistteleoperatedwholebodyimitation,allshire2025visual,araujo2025retargeting,lu2025mobile,sun2025spark}, and whole-body control~\cite{cheng2024expressive,feng2014optimization,henze2016passivity,herzog2014balancing,sentis2006whole}.

% Despite substantial progress in motion transfer, perception transfer remains comparatively underexplored in humanoid teleoperation. Although most systems provide egocentric visual feedback through onboard cameras, limitations in field of view, stereoscopic depth perception, viewpoint adaptation, and camera-motion coupling can still limit spatial perception and increase cognitive workload. Such limitations become particularly evident when operators must search for objects outside the current view, monitor multiple objects simultaneously, perform precise spatial alignment, or interact with dynamic elements in the environment. More broadly, long-horizon loco-manipulation requires operators to continuously gather, update, and utilize spatial information while maintaining task execution. When perceptual feedback becomes inefficient, operators' spatial perception degrades, reducing both operational efficiency and robustness. These challenges suggest that effective humanoid teleoperation depends not only on transferring human motion to the robot, but also on preserving the operator's spatial perception through effective spatial perception transfer.

However, reliable long-horizon teleoperation requires more than accurate motion transfer or retargeting. A central requirement in long-horizon loco-manipulation is that operators need to maintain task-relevant spatial awareness over time, including the locations of objects, the surrounding environment, robot's limb configurations, robot's pose relative to the scene, and so on. We refer to the spatial and temporal extent over which such awareness can be reliably maintained as the operator's \emph{perceptual horizon}. When such perception horizon is limited, the operator may still be able to issue motion commands, but task execution during the teleoperation becomes less efficient and less convenient: objects must be repeatedly searched for, recovery from unexpected events becomes slower, precise alignment becomes more difficult, etc. 

\begin{wrapfigure}{R}{0.5\textwidth}
    \centering
    \vspace{-10pt}
    \includegraphics[width=0.48\textwidth]{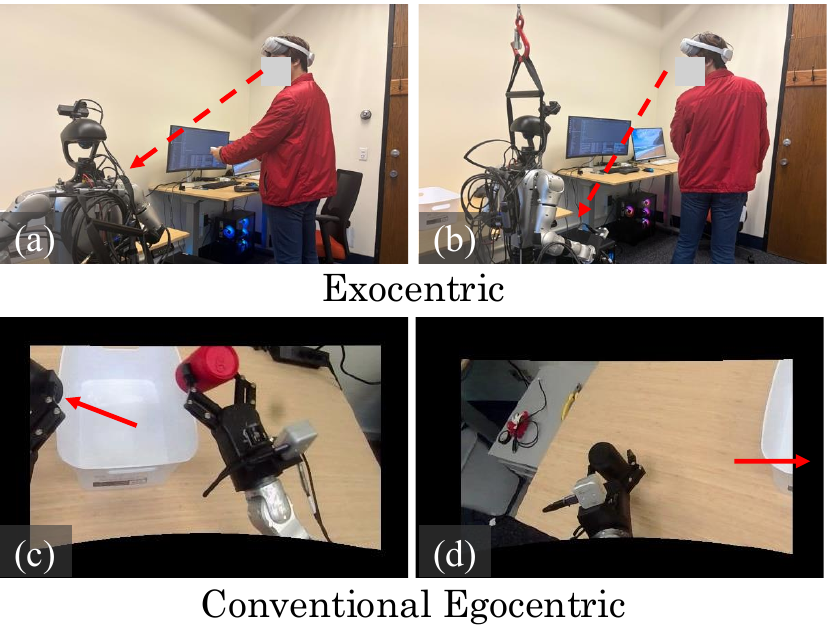}
    \vspace{-5pt}
    \caption{\textbf{Limitations of existing teleoperation interfaces.} In exocentric interfaces~\cite{he2024omnih2ouniversaldexteroushumantohumanoid,luo2026sonicsupersizingmotiontracking,ze2025twistteleoperatedwholebodyimitation,pan2025ams}, the separation between the operator and the robot (a) weakens depth perception and introduces additional difficulties when the operator turns (b). Conventional egocentric interfaces~\cite{ze2025twist2scalableportableholistic,li2025amoadaptivemotionoptimization,li2025cloneclosedloopwholebodyhumanoid} may hide manipulators (c) and lose track of objects outside the camera FOV (d).}
    \label{fig:existing}
    \vspace{-15pt}
\end{wrapfigure}

Existing egocentric teleoperation interfaces~\cite{ze2025twist2scalableportableholistic,li2025amoadaptivemotionoptimization,ben2025homiehumanoidlocomanipulationisomorphic,li2025cloneclosedloopwholebodyhumanoid} suffer from limited perceptual horizon in several ways. Narrow camera views may miss peripheral events that are important for task execution, such as objects falling outside the central view or robot limbs entering the workspace from the side. Robot-mounted cameras can also become unstable during locomotion, balancing, and recovery, making it difficult for the operator to maintain a consistent spatial reference. In addition, when visual exploration is coupled to robot actuation, head motion used for looking around may unintentionally command the robot's head, camera, or torso. These limitations are more pronounced in long-horizon teleoperation, where operators must continuously gather and update spatial information while simultaneously controlling locomotion and manipulation; See Fig.~\ref{fig:existing}.

In this work, we present \textbf{\name}~(\namelong), a perception-centered VR teleoperation system for long-horizon humanoid loco-manipulation. Unlike prior systems that primarily optimize motion retargeting~\cite{ze2025twistteleoperatedwholebodyimitation,araujo2025retargeting} and treat egocentric video as passive feedback~\cite{ben2025homiehumanoidlocomanipulationisomorphic,li2025cloneclosedloopwholebodyhumanoid}, \name~treats spatial perception as an explicit interface design objective. The system combines a robot-mounted binocular fisheye camera, a wide-field stereoscopic display, viewpoint-decoupled free-looking, and visual stabilization to provide a robot-centric view that is wide, stable, and actively inspectable. This design aims to extend the operator's perceptual horizon by preserving access to task-relevant spatial information over longer time horizons and larger operating ranges. A key design choice in \name~is to decouple visual exploration from robot actuation.
This differs from active-camera egocentric settings, such as TWIST2-style interfaces~\cite{ze2025twist2scalableportableholistic}, where the operator's head motion is used to change the robot-mounted camera viewpoint and visual exploration is therefore coupled to robot-side motion.
In contrast, \name~renders the egocentric stereo observation on a virtual hemisphere surrounding the operator, rather than rigidly attaching the view to the headset or directly mapping it to robot head, camera, or torso motion.
As a result, natural head rotations only change where the operator looks within the wide-field observation, without commanding the robot head, camera, or torso.
In parallel, visual stabilization compensates for camera rotations induced by robot motion, allowing the operator to perceive a more consistent spatial frame during locomotion and recovery.
% A key design choice in \name~is to decouple visual exploration from robot actuation. The egocentric stereo observation is rendered on a virtual hemisphere surrounding the operator, rather than being rigidly attached to the headset view or directly mapped to robot head or torso motion. As a result, natural head rotations change where the operator looks within the wide-field observation, without commanding the robot head, camera, or torso. In parallel, visual stabilization compensates for camera rotations induced by robot motion, allowing the operator to perceive a more consistent spatial frame during locomotion and recovery. Together, these components provide an observation interface that supports active visual search while preserving intuitive teleoperation control. \Lixing{
% Comments on ``decouple visual''.
% }

We evaluate \name~on a suite of perception-critical humanoid loco-manipulation tasks, including unexpected drop recovery, peripheral object retrieval, wide-range bimanual manipulation, fine spatial alignment, and dynamic interaction. These tasks are designed to stress different aspects of spatial perception, including object reacquisition, peripheral awareness, large-workspace coordination, and stable perception under robot motion. Compared with teleoperation paradigms inspired by recent humanoid data-collection systems, \name~improves task efficiency and accuracy, reduces search and recovery time. These results suggest that perception transfer is an important complement to motion transfer for scalable long-horizon humanoid teleoperation.

In summary, our contributions are threefold:
\begin{itemize}
    \item We formulate the operator's perceptual horizon as a practical bottleneck in long-horizon humanoid teleoperation, complementary to motion transfer.
    \item We introduce \name, a perception-centered VR teleoperation system with a wide-field, stereoscopic, stabilized, and viewpoint-decoupled robot-centric visual interface.
    \item We evaluate \name~on perception-critical humanoid loco-manipulation tasks and show that improving spatial perception improves task efficiency, accuracy, and recovery.
\end{itemize}

%===============================================================================

\section{Related Works}

\label{sec:related}
\paragraph{Bimanual teleoperation}
Teleoperation studies emerged in a bimanual setting, where two arms are attached steadily to a static or moving platform. Bimanual teleoperation systems like ~\citep{fu2024mobilealohalearningbimanual, kwon2025paprleplugandplayroboticlimb, wu2024gellogenerallowcostintuitive} build an isomorphic bi-arm structure under direct human operation and map the kinematics to the actuator arms. Other works use low-cost exoskeletons~\citep{fang2025airexo2scalinggeneralizablerobotic, yang2024acecrossplatformvisualexoskeletonslowcost} to collect human arm trajectories. Earlier VR/AR headsets also allowed relative hand position tracking, making room for VR-based bimanual teleoperation~\citep{zhang2018deepimitationlearningcomplex, PAN2021102167, ding2024bunnyvisionprorealtimebimanualdexterous}.

\paragraph{Humanoid teleoperation}
Humanoid robots' teleoperation imposes significantly greater challenges on accuracy and efficiency. Collecting~\citep{yin2025unitrackerlearninguniversalwholebody} and retargeting whole-body joints~\citep{darvish2019wholebodygeometricretargetinghumanoid, yang2025omniretargetinteractionpreservingdatageneration} is harder than bi-arm joints, as it's infeasible to have an isomorphic humanoid under the operator's direct control. ~\citep{ze2025twistteleoperatedwholebodyimitation, pan2025ams, xiong2026extremcontrollowlatencyhumanoidteleoperation, zhu2026clotclosedloopglobalmotion} relied on motion capturing (MoCap) systems to achieve the high-fidelity tracking necessary for whole-body control. ~\citep{ben2025homiehumanoidlocomanipulationisomorphic} uses exoskeleton interfaces and~\citep{li2026telegatewholebodyhumanoidteleoperation} adopts inertial motion capture equipment for whole-body teleoperation. More recently, VR has become the dominant interface for capturing operator motion ~\citep{ze2025twist2scalableportableholistic,cheng2024open,he2024omnih2ouniversaldexteroushumantohumanoid, luo2026sonicsupersizingmotiontracking, li2025amoadaptivemotionoptimization, li2025cloneclosedloopwholebodyhumanoid,allshire2025visual} due to its accessibility.

\paragraph{Immersive visual feedback and viewpoint control.}
Prior telepresence systems have used spherical re-rendering to reduce apparent latency while a robot-mounted stereo head follows operator motion~\citep{schwarz2021low}. Decoupled viewpoint control has also been studied in immersive humanoid teleoperation by combining live point-cloud observations with SLAM-reconstructed geometry outside the current camera view~\citep{chen2022enhanced}. Rotation compensation has been shown to improve comfort and reduce VR sickness in immersive telepresence~\citep{suomalainen2022unwinding}.
	
%===============================================================================

\section{Methods}
\label{sec:method}
\begin{wrapfigure}{R}{0.6\textwidth}
    \centering
    \vspace{-10pt}
    \includegraphics[width=0.6\textwidth]{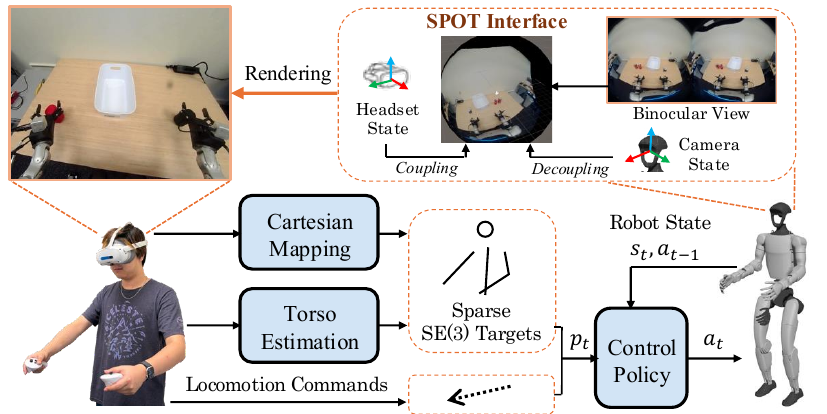}
    \caption{\textbf{Overview of the system architecture.}}
    \label{fig:pipeline}
    \vspace{-10pt}
\end{wrapfigure}

We present \name, a perception-centered teleoperation framework designed to enhance operator spatial perception during long-horizon humanoid loco-manipulation. The key design principle of \name~is to improve spatial perception transfer from the robot to the human operator, enabling efficient environmental understanding, object localization, and interaction while preserving intuitive teleoperation control. To achieve this goal, \name~combines a robot-centric immersive perception interface with a VR-based teleoperation pipeline. The perception interface provides binocular visual feedback through a stabilized wide-field stereoscopic display, allowing operators to freely explore the environment while maintaining a visually consistent representation of the world. In parallel, sparse observations from a VR headset and handheld controllers are converted into robot control targets and executed by a target-conditioned control policy. Figure~\ref{fig:pipeline} illustrates the overall system architecture.

\subsection{Immersive Perception Interface}
\label{sec:immersive_perception}

\paragraph{Egocentric Binocular View}
We provide the operator with an immersive egocentric observation of the robot environment through a robot-mounted binocular fisheye camera. The camera stream is received in Unity~\cite{unity} as a side-by-side (SBS) stereo video texture, and then rendered inside the VR headset through a hemispherical display surface. For each rendered viewing direction $\mathbf{d} \in \mathbb{S}^{2}$ in the front hemisphere, we project the direction into the corresponding fisheye image using an equisolid projection $
    r(\theta) =
    R \cdot \sin(\theta/2)/
    {\sin(\theta_{\max}/2)},
$
where $\theta = \arccos(d_z)$, $\theta_{\max}$ is the maximum visible polar angle, and $R$ is the normalized radius of the valid fisheye region. The left and right halves of the SBS image are sampled independently according to the rendered eye, preserving binocular disparity in the headset. Directions outside the forward-facing fisheye support are masked out. Compared with displaying a conventional planar camera image, this representation preserves wide peripheral coverage and binocular depth cues, which are particularly important in our tasks where relevant objects or robot limbs may enter the operator's peripheral field of view.

\paragraph{Viewpoint-Decoupled Observation}
The egocentric observation is rendered on a virtual hemisphere surrounding the operator rather than being rigidly attached to the operator's headset view. Consequently, the operator can naturally rotate their head to inspect different portions of the robot's wide-field observation without directly commanding a corresponding robot camera or head motion. This decouples visual exploration from robot actuation: the operator's head motion controls where they look within the currently available binocular observation, while robot commands are obtained separately from the tracked headset and handheld-controller signals. This design is especially suitable for our setting, where the robot does not require an actuated neck to provide an immersive observation interface.

\paragraph{Visual Stabilization}
Because the stereo camera is mounted on the moving robot, disturbances and recovery motions can induce abrupt camera rotations that make raw egocentric video difficult to interpret. To maintain a stable visual reference, we compensate for camera motion directly in the virtual display. Let $\mathbf{R}_t^{c}$ denote the camera orientation at time $t$ and $\mathbf{R}_0^{c}$ the orientation at initialization. The rendered observation is rotated according to the relative camera motion $
    \mathbf{R}_{t}^{\mathrm{rel}}
    =
    \left(\mathbf{R}_{0}^{c}\right)^{-1}
    \mathbf{R}_{t}^{c},
$
such that the rotational motion of the physical camera is counteracted in the displayed observation. Consequently, the operator perceives a visually stabilized scene despite camera movement induced by robot locomotion, balancing, or external interference.

The camera orientation estimate can be obtained from any source that provides the camera pose with respect to a global reference frame, such as robot state or an independent sensing device on the camera rig. In our implementation, we employ a lightweight IMU~\cite{wt61cimu} attached to the stereo camera and use its orientation estimate to drive the stabilization transform.

\subsection{Teleoperation Pipeline}
\label{sec:teleoperation_pipeline}

\paragraph{Human Motion Acquisition}
We acquire sparse human motion commands through an OpenXR-compatible VR headset and two handheld controllers. The interface is implemented using OpenXR~\cite{openxr} and is therefore hardware-agnostic across compliant VR platforms. We have validated the system on both Meta Quest~\cite{metaquest} and PICO~\cite{pico4} headsets without modifying the teleoperation pipeline.

In each Unity update cycle, the interface queries the tracked poses of the head-mounted display and the left and right controllers, yielding three rigid-body transforms:
$
    \mathcal{X}_{t}^{\mathrm{human}}
    =
    \left\{
        \mathbf{T}_{t}^{H},
        \mathbf{T}_{t}^{L},
        \mathbf{T}_{t}^{R}
    \right\},
$
where $H$, $L$, and $R$ denote the headset, left controller, and right controller, respectively. In addition to these poses, we read the two-dimensional joystick inputs and controller-button states. The acquired signals are serialized into a lightweight UDP message and streamed to the downstream teleoperation process at a target frequency of $50\sim90$~Hz. This interface requires only a headset and handheld controllers: the tracked head and hand poses provide the sparse upper-body motion specification, while joystick inputs are used to issue locomotion commands. In addition, the left and right controller triggers are directly mapped to the corresponding robot gripper commands.

\paragraph{Upper-body Estimation}
To decouple robot torso control from the operator's free head motion, we do not directly map the headset orientation to the robot torso. Instead, we estimate the operator's torso orientation from a short history of the three tracked VR poses using a lightweight streaming network. The network is trained on VR motion sequences with torso orientations recorded by a body-mounted IMU, while deployment requires only the headset and two handheld controllers. Given the predicted torso orientation, we recover the torso position from the tracked headset position using fixed headset-to-neck and neck-to-torso offsets. Together with the directly observed controller poses, this yields sparse upper-body targets consisting of the estimated torso and two hands. This estimation allows the operator to freely explore the environment through natural head motion while independently commanding robot body orientation.

\paragraph{Cartesian Retargeting}
Following the Cartesian-space formulation of ExtremControl~\cite{xiong2026extremcontrollowlatencyhumanoidteleoperation}, we directly map the estimated human upper-body poses to robot-side SE(3) targets without online inverse kinematics or joint-space retargeting. In contrast to the full-body setting, our interface maps only the torso and two wrists, while locomotion is commanded separately through the controller joystick. A one-shot neutral-pose calibration determines the initial heading alignment, controller-to-wrist rotational offsets, and human-side arm proportions. At runtime, each hand target is expressed in the torso frame and rescaled according to the corresponding human-to-robot arm-length ratio:
\begin{equation}
    \mathbf{p}_{t,k}^{r,\mathrm{rel}}
    =
    \mathbf{s}_{k}^{r}
    +
    \frac{l_{k}^{r}}{l_{k}^{h}}
    \left(
        \mathbf{p}_{t,k}^{h,\mathrm{rel}}
        -
        \mathbf{s}_{k}^{h}
    \right),
    \qquad k \in \{L,R\},
\end{equation}
where $\mathbf{s}_{k}$ denotes the shoulder anchor and $l_k$ denotes arm length. The relative hand orientations are transferred after applying the calibrated offsets, and the resulting torso and wrist poses are provided directly as Cartesian targets to the downstream control policy.

\paragraph{Target-conditioned Control Policy} Following ExtremControl~\cite{xiong2026extremcontrollowlatencyhumanoidteleoperation}, we train a target-conditioned policy in simulation that tracks Cartesian hand targets and locomotion commands. The policy receives robot proprioception, target commands, and recent action history, and is trained with standard motion-tracking objectives and domain randomization for deployment robustness.
	
%===============================================================================

\section{Experimental Results}
\label{sec:experiment}
\subsection{Experimental Setup}
\label{sec:experimental_setup}

\begin{figure}[t]
    \centering
    \captionsetup{type=figure}
    \includegraphics[width=0.95\textwidth]{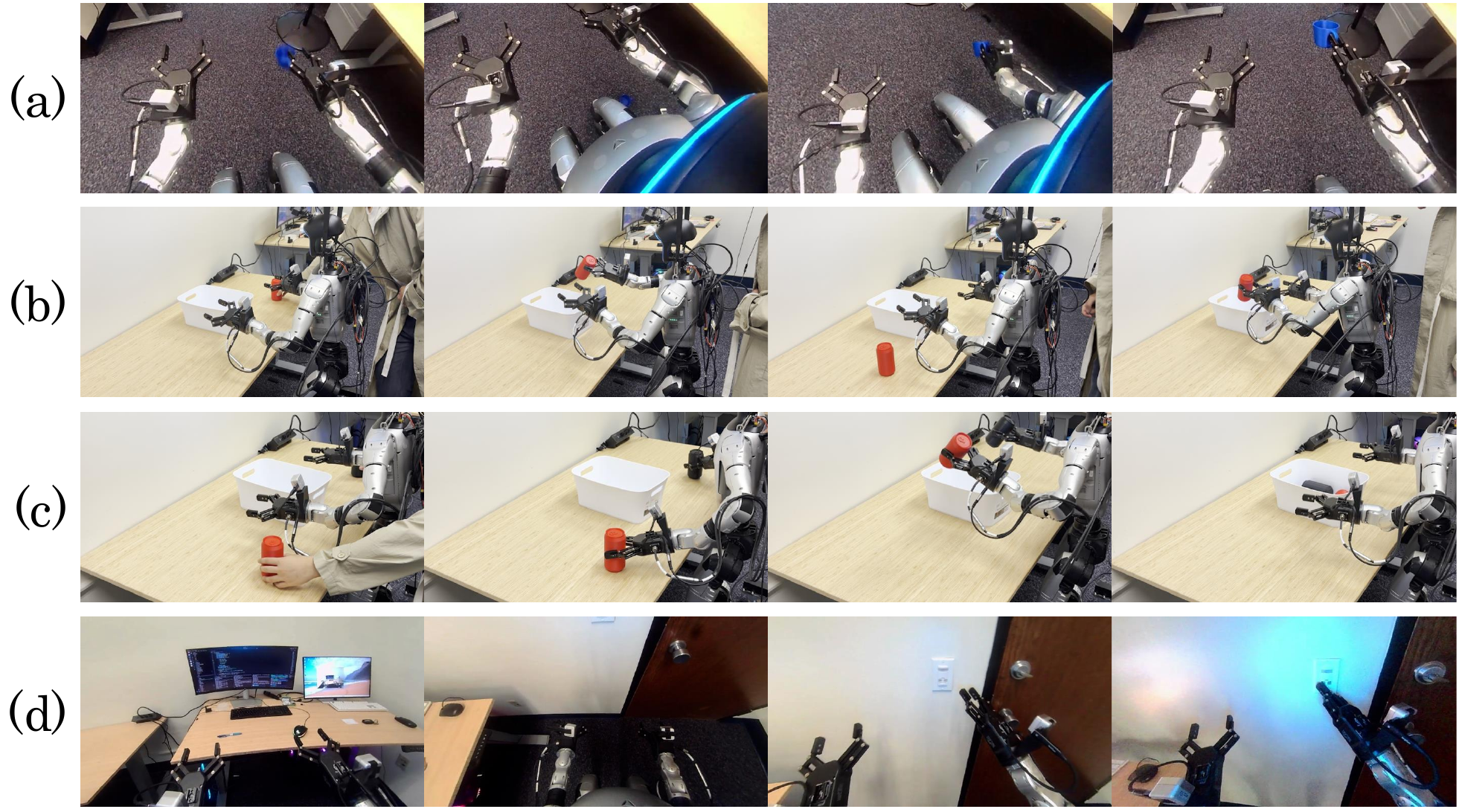}
    \caption{\textbf{Experimental task definitions.} (a) unexpected drop and recovery, (b) peripheral object retrieval, (c) multi-object bimanual retrieval, and (d) light switch.}
    \vspace{-10pt}
    \label{fig:task_definition}
\end{figure}

We evaluate our teleoperation interface on a set of long-horizon tasks that require continuous spatial awareness, object monitoring, and coordinated locomotion and manipulation.

We compare our method against these representative teleoperation baselines:

\begin{itemize}
    \item \textbf{Exocentric Teleoperation} which provides a third-person external observation similar to prior exocentric humanoid teleoperation systems such as OmniH2O~\cite{he2024omnih2ouniversaldexteroushumantohumanoid}, TWIST~\cite{ze2025twistteleoperatedwholebodyimitation}, and SONIC~\cite{luo2026sonicsupersizingmotiontracking}. In this setting, the operator controls the robot while observing the robot and workspace from an external viewpoint.
    \item \textbf{Conventional Egocentric Teleoperation} which uses a standard onboard stereo first-person camera view (e.g., ZED Mini~\cite{zedmini}) through the VR headset, including Open-Television~\cite{cheng2024open}, AMO~\cite{li2025amoadaptivemotionoptimization}, and TWIST2~\cite{ze2025twist2scalableportableholistic}. In this setting, we allow the operator to freely rotate the viewing direction through head motion, and simulate the corresponding viewpoint changes similar to their robot-mounted active camera setup. However, the observable region remains constrained by the limited onboard camera field of view. Specifically, it renders a $110^\circ \times 70^\circ$ viewport with the same shader as in TWIST2 after an inverse equisolid projection, with HMD yaw/pitch mapped to viewport rotation within the fisheye image. It does not physically actuate a camera and its modeled rotation range is $160^\circ$ with no extra delay. This serves as a matched active-view egocentric baseline inspired by prior systems, not a full reproduction of their hardware. 
\end{itemize}

We report mean task completion time $\pm$ SD over successful trials across all methods performed by 10 operators (3 experienced and 7 novice). Each participant completed 20 trials per task per interface. To ensure fair comparison, all methods share the same robot platform, teleoperation pipeline, task environments, and protocols; only the perception interface differs. The interface order was randomized across participants.

We evaluate four quantitative tasks covering different spatial perception challenges:

\begin{itemize}
    \item \textbf{Unexpected Drop and Recovery}, as shown in Fig.~\ref{fig:task_definition}(a). The robot transports an object that is unexpectedly dropped during locomotion. The operator must rapidly localize, recover, and continue the task. We additionally measure \textbf{object recovery time} to quantify recovery efficiency after the unexpected drop.
    \item \textbf{Peripheral Object Retrieval}, as shown in Fig.~\ref{fig:task_definition}(b). A target object randomly appears outside the central viewing region. The operator must detect and retrieve it as quickly as possible. We additionally measure \textbf{visual search time} to quantify the time required to locate the target.
    \item \textbf{Multi-object Bimanual Retrieval}, as shown in Fig.~\ref{fig:task_definition}(c). Two target objects appear simultaneously on opposite sides of the robot. Operators must retrieve both while continuously monitoring both manipulators and task-relevant objects. We additionally record the number of \textbf{grasp attempts} to quantify manipulation accuracy during bimanual interaction.
    \item \textbf{Light Switch}, as shown in Fig.~\ref{fig:task_definition}(d). The robot starts facing away from a wall-mounted light switch and must locomote toward and precisely toggle it. We additionally measure \textbf{viewpoint realignment time} to quantify the visual reorientation required after robot turning.
\end{itemize}

\begin{table}[t]
    \centering
    \begin{tabular}{lccc}
    \toprule
    Method & Unexpected Drop $\downarrow$ & Recovery Time $\downarrow$ & Success Rate $\uparrow$ 
    \\
    \hline
    Exo & 25.73 $\pm$ 5.48\,s & 15.07 $\pm$ 7.80\,s & 79.5 $\%$ % Experienced 55, Novice 104
    \\
    \hline
    Ego & 32.12 $\pm$ 10.03\,s & 20.28 $\pm$ 8.06\,s & 96.5 $\%$ % Experienced 60, Novice 133
    \\
    \hline
    SPOT & \textbf{19.69 $\pm$ 4.04\,s } & \textbf{11.12 $\pm$ 3.69\,s } & \textbf{98 $\%$} % Experienced 60, Novice 136
    \\
    \midrule
     & Peripheral Retrieval $\downarrow$ & Search Time $\downarrow$ & Success Rate $\uparrow$ 
    \\
    \hline
    Exo & 9.18 $\pm$ 2.68\,s & \textbf{0.00 $\pm$ 0.00}\,s & 99 $\%$  % Experienced 60, Novice 138
    \\
    \hline
    Ego & 11.16 $\pm$ 2.04\,s & 0.96 $\pm$ 0.29\,s & \textbf{100 $\%$} % Experienced 60, Novice 140
    \\
    \hline
    SPOT & \textbf{7.43 $\pm$ 1.49\,s } & 0.11 $\pm$ 0.03\,s & \textbf{100 $\%$} % Experienced 60, Novice 140
    \\
    \midrule
     & Bimanual Retrieval $\downarrow$ & Grasp Attempts $\downarrow$ & Success Rate $\uparrow$ 
    \\
    \hline
    Exo & 13.64 $\pm$ 4.47\,s & 3.11 $\pm$ 0.74 & 93.5 $\%$ % Experienced 60, Novice 127
    \\
    \hline
    Ego & 15.45 $\pm$ 3.67\,s & 2.56 $\pm$ 0.58 & 99 $\%$ % Experienced 60, Novice 138
    \\
    \hline
    SPOT & \textbf{9.76 $\pm$ 2.82\,s } & \textbf{2.11 $\pm$ 0.33} & \textbf{99.5 $\%$} % Experienced 60, Novice 139
    \\
    \midrule
     & Light Switch $\downarrow$ & Alignment Time $\downarrow$ & Success Rate $\uparrow$ 
    \\
    \hline
    Exo & 16.86 $\pm$ 3.83\,s & N/A & 8 $\%$ % Experienced 16, Novice 0
    \\
    \hline
    Ego & 18.15 $\pm$ 4.45\,s & 1.52 $\pm$ 0.41\,s & \textbf{100 $\%$} % Experienced 60, Novice 140
    \\
    \hline
    SPOT & \textbf{11.54 $\pm$ 3.39\,s } & \textbf{0.39 $\pm$ 0.08\,s } & \textbf{100 $\%$} % Experienced 60, Novice 140
    \\[-2pt]
    \bottomrule
    \end{tabular}
    \vspace{10pt}
    \caption{\textbf{Quantitative evaluation across four tasks.} Our interface improves efficiency on recovery, peripheral awareness, bimanual coordination, and long-horizon loco-manipulation tasks.}
    \label{tab:task_time}
\end{table}

\subsection{Awareness of Unexpected Events and Out-of-View Objects}

We evaluate this capability using the Unexpected Drop and Recovery task and the Peripheral Object Retrieval task. Tab.~\ref{tab:task_time} reports the completion times of these tasks.

\begin{wrapfigure}{R}{0.48\textwidth}
    \centering
    \vspace{-15pt}
    \includegraphics[width=0.48\textwidth]{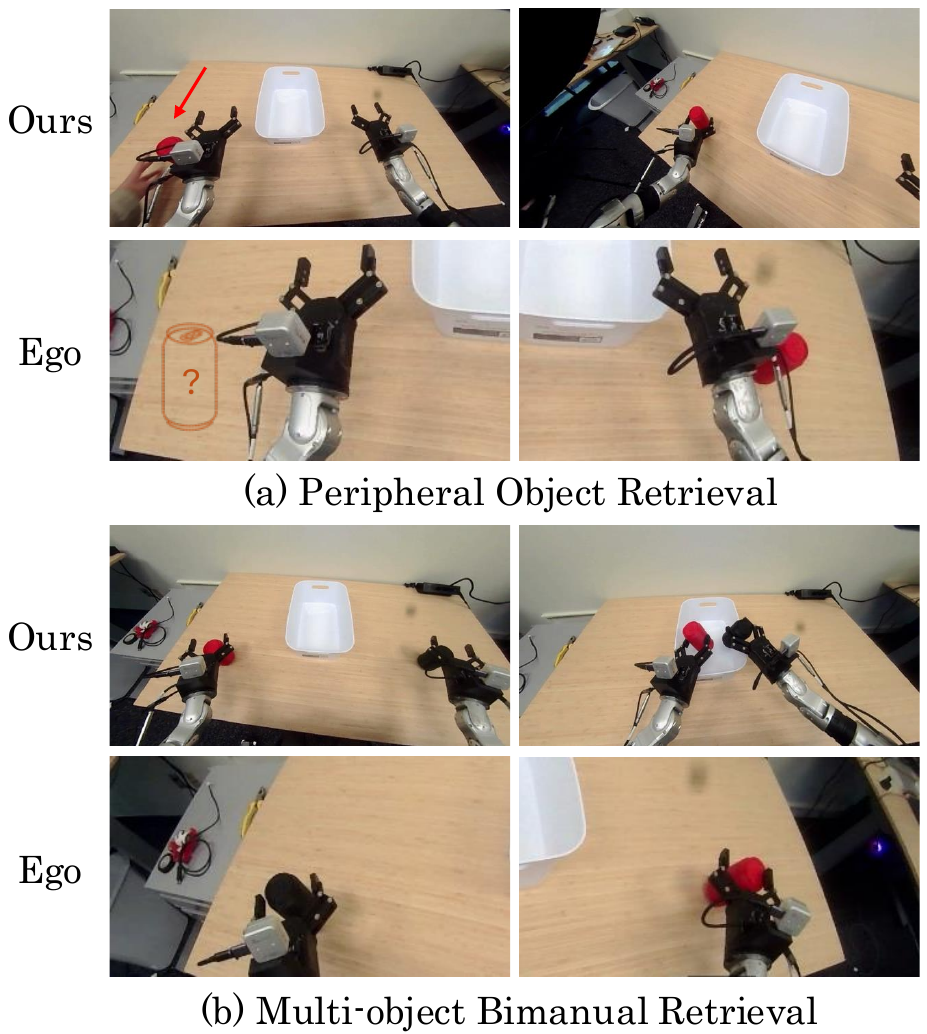}
    \caption{\textbf{Comparison between our interface (Ours) and conventional egocentric teleoperation (Ego).} The limited FOV in conventional egocentric interfaces (a) requires additional search and re-localization time, and (b) requires sequential instead of simultaneous manipulation.}
    \label{fig:ego_comparison}
    \vspace{-25pt}
\end{wrapfigure}

The exocentric baseline preserves global scene visibility but suffers from reduced spatial precision during manipulation, especially when the manipulating space is not as ideal as a tabletop setup. In contrast, the task-specific search-time metric clarifies that the conventional egocentric baseline provides immersive first-person feedback but loses awareness of objects once they leave the narrow onboard camera field of view. It suffers from difficulty locating out-of-view objects when they come into the manipulation workspace, and requires extra search time, as shown in Fig.~\ref{fig:ego_comparison}(a).

SPOT reduces recovery time compared with both baselines and substantially reduces visual search time compared with Conventional Egocentric teleoperation, showing faster recovery from unexpected object loss. The immersive wide-field interface enables operators to rapidly re-localize dropped or peripheral objects while maintaining robot-centered spatial consistency throughout the task.

We also show in Fig.~\ref{fig:teaser}(b) that the operator can dynamically locate a ping-pong ball during teleoperation and return it to the opponent in real time. This demonstrates that our interface is capable of tracking dynamic objects across a large field of view.

\subsection{Bimanual Spatial Awareness}

We evaluate the ability to simultaneously monitor both manipulators and multiple task-relevant objects during coordinated bimanual interaction by using the Multi-object Bimanual Retrieval task. Tab.~\ref{tab:task_time} summarizes the quantitative results.

The conventional egocentric baseline loses visibility of one manipulator when the operator focuses on the other hand, particularly during large-arm motions or object transfer. The operator needs to switch focus and complete the task on each manipulator one at a time, as shown in Fig.\ref{fig:ego_comparison}. Meanwhile, although the exocentric baseline preserves visibility of both arms, the detached external viewpoint weakens depth perception and precise spatial judgment during manipulation. In our experiments, the operator also experiences reduced manipulation accuracy on the arm farther from the viewing position.

Our method improves the task efficiency while reducing grasp attempts by enabling continuous awareness of both manipulators and nearby objects within a coherent immersive spatial representation.

We further present several challenging qualitative demonstrations that highlight additional bimanual capabilities. Fig.~\ref{fig:teaser}(a) shows the robot performing rope knotting while continuously monitoring both manipulators and rope configuration, highlighting long-horizon bimanual spatial awareness. Fig.~\ref{fig:teaser}(d) shows the robot inserting a toy sword into a sheath during large-range dual-arm motions, requiring simultaneous global spatial awareness and precise bimanual alignment.

\subsection{Spatial Consistency Across Locomotion and Manipulation}

We further evaluate spatial perception during transitions between navigation and precise manipulation using the Light Switch task. This task requires operators to maintain environmental awareness during locomotion while also performing accurate final-stage interaction. Tab.~\ref{tab:task_time} shows the quantitative comparison.

\begin{wrapfigure}{R}{0.7\textwidth}
    \centering
    \vspace{-10pt}
    \includegraphics[width=0.7\textwidth]{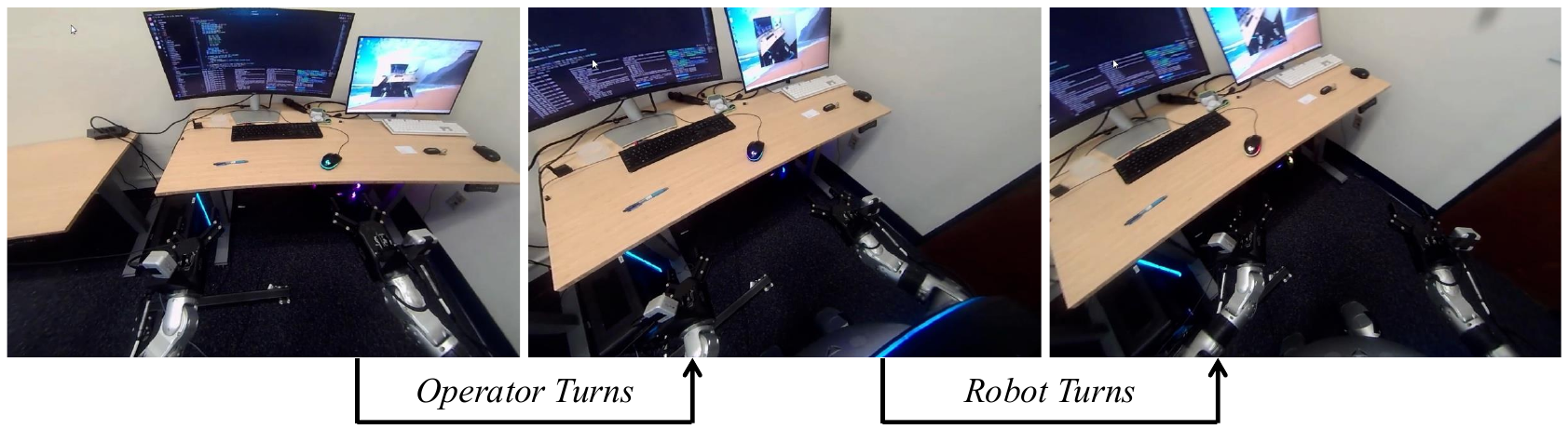}
    \caption{\textbf{Stabilized egocentric perception during turning motions.} When the operator turns in VR, their view turns accordingly. When the robot turns while the operator stops, the coordinate frame in VR remains stabilized, preserving spatial continuity and reducing abrupt viewpoint changes.}
    \label{fig:stabilization}
    \vspace{-10pt}
\end{wrapfigure}

The exocentric baseline frequently fails because the operator loses awareness of the robot after reorientation. Conventional egocentric teleoperation preserves immersion but requires additional viewpoint realignment after robot turning, since the visual frame rotates together with the robot during navigation.

Our method maintains stable robot-centered spatial perception across both locomotion and manipulation stages. As shown in Fig.~\ref{fig:stabilization}, the viewpoint in VR is decoupled from robot motion and follows the head motion solely, avoiding abrupt viewpoint changes during teleoperation, which results in less viewpoint alignment time and improves task completion efficiency and reliability.

% \fbox{\parbox{\textwidth}{

% Experiment Tasks:

% 1. Unexpected Drop \& Recovery (Main)

% 2. Peripheral Object Retrieval (Beyond Current FOV)
 
% 3. Multi-object Bimanual Retrieval (Multi-Object)

% 4. Light Switch (Loco, Accuracy)

% Demo-Only Tasks:

% 5. Ping-pong (Dynamic)

% 6. Knot (Bimanual)

% 7. Sword and Sheath (Bimanual)

% }}

%===============================================================================

% \section{Conclusion}
% \label{sec:conclusion}
% \input{src/5-conclusion}

%===============================================================================

%===============================================================================
\section{Conclusion}
\label{sec:conclusion}
% \section{Conclusion}
We presented \name, a spatial perception-oriented VR teleoperation system for long-horizon humanoid data collection. Motivated by the observation that effective humanoid teleoperation requires not only accurate motion transfer but also sustained task-relevant spatial awareness, we introduced the notion of the operator's \emph{perceptual horizon}. \name extends this perceptual horizon through a wide-field stereoscopic robot-centric interface that combines binocular fisheye sensing, viewpoint-decoupled free-looking, and visual stabilization. By decoupling visual exploration from robot actuation, \name allows operators to actively inspect the environment without interfering with robot motion, enabling more flexible and user-friendly control during long-horizon loco-manipulation. Across perception-critical humanoid tasks involving drop recovery, peripheral retrieval, large-workspace bimanual manipulation, fine alignment, and dynamic interaction, \name improves task efficiency, accuracy, and recovery speed, compared with conventional egocentric teleoperation interfaces. These results indicate that perception transfer is a key complement to motion transfer for scalable humanoid data collection. More broadly, \name highlights spatial perception as an explicit interface design objective for teleoperation: improving how operators perceive, search, and maintain task-relevant spatial context can improve the quality, consistency, and scalability of long-horizon humanoid demonstrations.

% Our current system focuses primarily on immersive visual feedback; incorporating tactile, force, and audio feedback could further improve teleoperation in contact-rich settings. Finally, end-to-end visual streaming latency remains a bottleneck for highly responsive whole-body operation, motivating future work on lower-latency perception and rendering pipelines.

\section{Limitations and Future Work}
\label{sec:limitation}
% \noindent \textit{Limitations and Future Work.} % \paragraph{Large-scale Downstream Learning}
% While our experiments demonstrate that the proposed interface improves spatial awareness and teleoperation performance across a range of loco-manipulation tasks, we have not yet performed a large-scale study on downstream policy learning using massive collected demonstrations. The relationship between improved operator perception and the scalability of learned robot policies remains an important direction for investigation. % \paragraph{Limited Multi-Modal Perception}
Our current system primarily focuses on visual perception through immersive stereo observation. However, effective teleoperation also depends on other sensing modalities, including tactile feedback, force interaction, and audio. Incorporating additional sensory channels into immersive humanoid teleoperation remains an important future direction. % \paragraph{Streaming Latency}
Although the proposed interface improves operator spatial awareness, end-to-end camera streaming latency remains a major bottleneck for responsive teleoperation. Delays introduced by image transmission can reduce responsiveness during rapid manipulation or recovery behaviors. While our system remains usable under current latency conditions, further reducing visual feedback latency will likely be critical for highly responsive whole-body teleoperation.
While \name improves spatial awareness and teleoperation performance across loco-manipulation tasks, we have not yet evaluated its impact on large-scale downstream policy learning. Understanding how improved operator perception affects the scalability and quality of learned policies remains an important direction. 

\clearpage
% The acknowledgments are automatically included only in the final and preprint versions of the paper.
% \acknowledgments{}

%===============================================================================

\bibliography{references}  % .bib

@article{ze2025twistteleoperatedwholebodyimitation,
  title={Twist: Teleoperated whole-body imitation system},
  author={Ze, Yanjie and Chen, Zixuan and Ara{\'u}jo, Joao Pedro and Cao, Zi-ang and Peng, Xue Bin and Wu, Jiajun and Liu, C Karen},
  journal={arXiv preprint arXiv:2505.02833},
  year={2025}
}

@article{pan2025ams,
  title={Agility Meets Stability: Versatile Humanoid Control with Heterogeneous Data},
  author={Pan, Yixuan and Qiao, Ruoyi and Chen, Li and Chitta, Kashyap and Pan, Liang and Mai, Haoguang and Bu, Qingwen and Zheng, Cunyuan and Zhao, Hao and Luo, Ping and Li, Hongyang},
  journal={arXiv preprint arXiv:2511.17373},
  year={2025}
}

@article{xiong2026extremcontrollowlatencyhumanoidteleoperation,
  title={ExtremControl: Low-Latency Humanoid Teleoperation with Direct Extremity Control},
  author={Xiong, Ziyan and Fang, Lixing and Huang, Junyun and Yamazaki, Kashu and Zhang, Hao and Gan, Chuang},
  journal={arXiv preprint arXiv:2602.11321},
  year={2026}
}

@article{ben2025homiehumanoidlocomanipulationisomorphic,
  title={Homie: Humanoid loco-manipulation with isomorphic exoskeleton cockpit},
  author={Ben, Qingwei and Jia, Feiyu and Zeng, Jia and Dong, Junting and Lin, Dahua and Pang, Jiangmiao},
  journal={arXiv preprint arXiv:2502.13013},
  year={2025}
}

@article{he2024omnih2ouniversaldexteroushumantohumanoid,
  title={Omnih2o: Universal and dexterous human-to-humanoid whole-body teleoperation and learning},
  author={He, Tairan and Luo, Zhengyi and He, Xialin and Xiao, Wenli and Zhang, Chong and Zhang, Weinan and Kitani, Kris and Liu, Changliu and Shi, Guanya},
  journal={arXiv preprint arXiv:2406.08858},
  year={2024}
}

@article{luo2026sonicsupersizingmotiontracking,
  title={Sonic: Supersizing motion tracking for natural humanoid whole-body control},
  author={Luo, Zhengyi and Yuan, Ye and Wang, Tingwu and Li, Chenran and Chen, Sirui and Castaneda, Fernando and Cao, Zi-Ang and Li, Jiefeng and Minor, David and Ben, Qingwei and others},
  journal={arXiv preprint arXiv:2511.07820},
  year={2025}
}

@inproceedings{li2025cloneclosedloopwholebodyhumanoid,
  title={Clone: Closed-loop whole-body humanoid teleoperation for long-horizon tasks},
  author={Li, Yixuan and Lin, Yutang and Cui, Jieming and Liu, Tengyu and Liang, Wei and Zhu, Yixin and Huang, Siyuan},
  booktitle={9th Annual Conference on Robot Learning},
  year={2025}
}

@article{li2025amoadaptivemotionoptimization,
  title={Amo: Adaptive motion optimization for hyper-dexterous humanoid whole-body control},
  author={Li, Jialong and Cheng, Xuxin and Huang, Tianshu and Yang, Shiqi and Qiu, Ri-Zhao and Wang, Xiaolong},
  journal={arXiv preprint arXiv:2505.03738},
  year={2025}
}

@article{ze2025twist2scalableportableholistic,
  title={Twist2: Scalable, portable, and holistic humanoid data collection system},
  author={Ze, Yanjie and Zhao, Siheng and Wang, Weizhuo and Kanazawa, Angjoo and Duan, Rocky and Abbeel, Pieter and Shi, Guanya and Wu, Jiajun and Liu, C Karen},
  journal={arXiv preprint arXiv:2511.02832},
  year={2025}
}

@misc{metaquest,
  author = {{Meta.}},
  title = {Meta Quest VR Headsets and Accessories | Meta Store},
  year = 2026,
  url = {https://www.meta.com/quest/},
  urldate = {2026-05-28}
}

@misc{pico4,
  author = {{PICO Immersive Pte.ltd.}},
  title = {PICO 4 Ultra Enterprise-VR-MR-headset | PICO Global},
  year = 2026,
  url = {https://www.picoxr.com/global/products/pico4-ultra-enterprise},
  urldate = {2026-05-28}
}

@misc{zedmini,
  author = {{Stereolabs Inc.}},
  title = {Introducing ZED Mini | Stereolabs},
  year = 2026,
  url = {https://www.stereolabs.com/blog/introducing-zed-mini},
  urldate = {2026-05-28}
}

@article{cheng2024open,
  title={Open-television: Teleoperation with immersive active visual feedback},
  author={Cheng, Xuxin and Li, Jialong and Yang, Shiqi and Yang, Ge and Wang, Xiaolong},
  journal={arXiv preprint arXiv:2407.01512},
  year={2024}
}

@misc{unity,
  author = {{Unity Technologies}},
  title = {Unity},
  year = {2026},
  url = {https://unity.com/},
  urldate = {2026-05-28}
}

@misc{openxr,
  author = {{The Khronos Group Inc.}},
  title = {Khronos OpenXR Registry},
  year = {2026},
  url = {https://registry.khronos.org/OpenXR/},
  urldate = {2026-05-28}
}

@misc{wt61cimu,
  author = {{WitMotion HK Co.ltd.}},
  title = {6-axis accelerometer: Precision in Motion Tracking},
  year = {2026},
  url = {https://witmotion-sensor.com/},
  urldate = {2026-05-28}
}

@article{fu2024mobilealohalearningbimanual,
  title={Mobile aloha: Learning bimanual mobile manipulation with low-cost whole-body teleoperation},
  author={Fu, Zipeng and Zhao, Tony Z and Finn, Chelsea},
  journal={arXiv preprint arXiv:2401.02117},
  year={2024}
}

@article{kwon2025paprleplugandplayroboticlimb,
  title={PAPRLE (Plug-And-Play Robotic Limb Environment): A Modular Ecosystem for Robotic Limbs},
  author={Kwon, Obin and Yamsani, Sankalp and Myers, Noboru and Taylor, Sean and Hong, Jooyoung and Park, Kyungseo and Alspach, Alex and Kim, Joohyung},
  journal={arXiv preprint arXiv:2507.05555},
  year={2025}
}

@article{fang2025airexo2scalinggeneralizablerobotic,
  title={Airexo-2: Scaling up generalizable robotic imitation learning with low-cost exoskeletons},
  author={Fang, Hongjie and Wang, Chenxi and Wang, Yiming and Chen, Jingjing and Xia, Shangning and Lv, Jun and He, Zihao and Yi, Xiyan and Guo, Yunhan and Zhan, Xinyu and others},
  journal={arXiv preprint arXiv:2503.03081},
  year={2025}
}

@inproceedings{darvish2019wholebodygeometricretargetinghumanoid,
  title={Whole-body geometric retargeting for humanoid robots},
  author={Darvish, Kourosh and Tirupachuri, Yeshasvi and Romualdi, Giulio and Rapetti, Lorenzo and Ferigo, Diego and Chavez, Francisco Javier Andrade and Pucci, Daniele},
  booktitle={2019 IEEE-RAS 19th International Conference on Humanoid Robots (Humanoids)},
  pages={679--686},
  year={2019},
  organization={IEEE}
}

@article{yang2025omniretargetinteractionpreservingdatageneration,
  title={Omniretarget: Interaction-preserving data generation for humanoid whole-body loco-manipulation and scene interaction},
  author={Yang, Lujie and Huang, Xiaoyu and Wu, Zhen and Kanazawa, Angjoo and Abbeel, Pieter and Sferrazza, Carmelo and Liu, C Karen and Duan, Rocky and Shi, Guanya},
  journal={arXiv preprint arXiv:2509.26633},
  year={2025}
}

@article{yin2025unitrackerlearninguniversalwholebody,
  title={Unitracker: Learning universal whole-body motion tracker for humanoid robots},
  author={Yin, Kangning and Zeng, Weishuai and Fan, Ke and Dai, Minyue and Wang, Zirui and Zhang, Qiang and Tian, Zheng and Wang, Jingbo and Pang, Jiangmiao and Zhang, Weinan},
  journal={IEEE Robotics and Automation Letters},
  year={2026},
  publisher={IEEE}
}

@article{li2026telegatewholebodyhumanoidteleoperation,
  title={TeleGate: Whole-Body Humanoid Teleoperation via Gated Expert Selection with Motion Prior},
  author={Li, Jie and Tang, Bing and Wu, Feng},
  journal={arXiv preprint arXiv:2602.09628},
  year={2026}
}

@article{zhu2026clotclosedloopglobalmotion,
  title={Clot: Closed-loop global motion tracking for whole-body humanoid teleoperation},
  author={Zhu, Tengjie and Cai, Guanyu and Zhaohui, Yang and Ren, Guanzhu and Xie, Haohui and Wang, ZiRui and Wu, Junsong and Wang, Jingbo and Yang, Xiaokang and Mu, Yao and others},
  journal={arXiv preprint arXiv:2602.15060},
  year={2026}
}

@inproceedings{he2024learning,
  title={Learning human-to-humanoid real-time whole-body teleoperation},
  author={He, Tairan and Luo, Zhengyi and Xiao, Wenli and Zhang, Chong and Kitani, Kris and Liu, Changliu and Shi, Guanya},
  booktitle={2024 IEEE/RSJ International Conference on Intelligent Robots and Systems (IROS)},
  pages={8944--8951},
  year={2024},
  organization={IEEE}
}

@article{allshire2025visual,
  title={Visual imitation enables contextual humanoid control},
  author={Allshire, Arthur and Choi, Hongsuk and Zhang, Junyi and McAllister, David and Zhang, Anthony and Kim, Chung Min and Darrell, Trevor and Abbeel, Pieter and Malik, Jitendra and Kanazawa, Angjoo},
  journal={arXiv preprint arXiv:2505.03729},
  year={2025}
}

@article{araujo2025retargeting,
  title={Retargeting matters: General motion retargeting for humanoid motion tracking},
  author={Araujo, Joao Pedro and Ze, Yanjie and Xu, Pei and Wu, Jiajun and Liu, C Karen},
  journal={arXiv preprint arXiv:2510.02252},
  year={2025}
}

@inproceedings{lu2025mobile,
  title={Mobile-television: Predictive motion priors for humanoid whole-body control},
  author={Lu, Chenhao and Cheng, Xuxin and Li, Jialong and Yang, Shiqi and Ji, Mazeyu and Yuan, Chengjing and Yang, Ge and Yi, Sha and Wang, Xiaolong},
  booktitle={2025 IEEE International Conference on Robotics and Automation (ICRA)},
  pages={5364--5371},
  year={2025},
  organization={IEEE}
}

@inproceedings{sun2025spark,
  title={SPARK: Safe protective and assistive robot kit},
  author={Sun, Yifan and Chen, Rui and Yun, Kai S and Fang, Yikuan and Jung, Sebin and Li, Feihan and Li, Bowei and Zhao, Weiye and Liu, Changliu},
  year={2025},
  organization={IFAC Symposium on Robotics}
}

@inproceedings{myers2025child,
  title={CHILD (Controller for Humanoid Imitation and Live Demonstration): A Whole-Body Humanoid Teleoperation System},
  author={Myers, Noboru and Kwon, Obin and Yamsani, Sankalp and Kim, Joohyung},
  booktitle={2025 IEEE-RAS 24th International Conference on Humanoid Robots (Humanoids)},
  pages={1--6},
  year={2025},
  organization={IEEE}
}

@inproceedings{purushottam2025heavy,
  title={Heavy lifting tasks via haptic teleoperation of a wheeled humanoid},
  author={Purushottam, Amartya and Yan, Jack and Xu, Christopher and Ramos, Joao},
  booktitle={2025 IEEE-RAS 24th International Conference on Humanoid Robots (Humanoids)},
  pages={345--350},
  year={2025},
  organization={IEEE}
}

@article{cheng2024expressive,
  title={Expressive whole-body control for humanoid robots},
  author={Cheng, Xuxin and Ji, Yandong and Chen, Junming and Yang, Ruihan and Yang, Ge and Wang, Xiaolong},
  journal={arXiv preprint arXiv:2402.16796},
  year={2024}
}

@inproceedings{feng2014optimization,
  title={Optimization based full body control for the atlas robot},
  author={Feng, Siyuan and Whitman, Eric and Xinjilefu, X and Atkeson, Christopher G},
  booktitle={2014 IEEE-RAS International Conference on Humanoid Robots},
  pages={120--127},
  year={2014},
  organization={IEEE}
}

@article{henze2016passivity,
  title={Passivity-based whole-body balancing for torque-controlled humanoid robots in multi-contact scenarios},
  author={Henze, Bernd and Roa, Maximo A and Ott, Christian},
  journal={The International Journal of Robotics Research},
  volume={35},
  number={12},
  pages={1522--1543},
  year={2016},
  publisher={SAGE Publications Sage UK: London, England}
}

@inproceedings{herzog2014balancing,
  title={Balancing experiments on a torque-controlled humanoid with hierarchical inverse dynamics},
  author={Herzog, Alexander and Righetti, Ludovic and Grimminger, Felix and Pastor, Peter and Schaal, Stefan},
  booktitle={2014 IEEE/RSJ International Conference on Intelligent Robots and Systems},
  pages={981--988},
  year={2014},
  organization={IEEE}
}

@inproceedings{sentis2006whole,
  title={A whole-body control framework for humanoids operating in human environments},
  author={Sentis, Luis and Khatib, Oussama},
  booktitle={Proceedings 2006 IEEE International Conference on Robotics and Automation, 2006. ICRA 2006.},
  pages={2641--2648},
  year={2006},
  organization={IEEE}
}

@inproceedings{wu2024gellogenerallowcostintuitive,
  title={Gello: A general, low-cost, and intuitive teleoperation framework for robot manipulators},
  author={Wu, Philipp and Shentu, Yide and Yi, Zhongke and Lin, Xingyu and Abbeel, Pieter},
  booktitle={2024 IEEE/RSJ International Conference on Intelligent Robots and Systems (IROS)},
  pages={12156--12163},
  year={2024},
  organization={IEEE}
}

@article{yang2024acecrossplatformvisualexoskeletonslowcost,
  title={Ace: A cross-platform visual-exoskeletons system for low-cost dexterous teleoperation},
  author={Yang, Shiqi and Liu, Minghuan and Qin, Yuzhe and Ding, Runyu and Li, Jialong and Cheng, Xuxin and Yang, Ruihan and Yi, Sha and Wang, Xiaolong},
  journal={arXiv preprint arXiv:2408.11805},
  year={2024}
}

@inproceedings{zhang2018deepimitationlearningcomplex,
  title={Deep imitation learning for complex manipulation tasks from virtual reality teleoperation},
  author={Zhang, Tianhao and McCarthy, Zoe and Jow, Owen and Lee, Dennis and Chen, Xi and Goldberg, Ken and Abbeel, Pieter},
  booktitle={2018 IEEE international conference on robotics and automation (ICRA)},
  pages={5628--5635},
  year={2018},
  organization={Ieee}
}

@article{PAN2021102167,
  title={Augmented reality-based robot teleoperation system using RGB-D imaging and attitude teaching device},
  author={Pan, Yong and Chen, Chengjun and Li, Dongnian and Zhao, Zhengxu and Hong, Jun},
  journal={Robotics and Computer-Integrated Manufacturing},
  volume={71},
  pages={102167},
  year={2021},
  publisher={Elsevier}
}

@inproceedings{ding2024bunnyvisionprorealtimebimanualdexterous,
  title={Bunny-visionpro: Real-time bimanual dexterous teleoperation for imitation learning},
  author={Ding, Runyu and Qin, Yuzhe and Zhu, Jiyue and Jia, Chengzhe and Yang, Shiqi and Yang, Ruihan and Qi, Xiaojuan and Wang, Xiaolong},
  booktitle={2025 IEEE/RSJ International Conference on Intelligent Robots and Systems (IROS)},
  pages={12248--12255},
  year={2025},
  organization={IEEE}
}

@inproceedings{schwarz2021low,
  title={Low-latency immersive 6D televisualization with spherical rendering},
  author={Schwarz, Max and Behnke, Sven},
  booktitle={2020 IEEE-RAS 20th International Conference on Humanoid Robots (Humanoids)},
  pages={320--325},
  year={2021},
  organization={IEEE}
}

@inproceedings{chen2022enhanced,
  title={Enhanced visual feedback with decoupled viewpoint control in immersive humanoid robot teleoperation using SLAM},
  author={Chen, Yang and Sun, Leyuan and Benallegue, Mehdi and Cisneros-Lim{\'o}n, Rafael and Singh, Rohan P and Kaneko, Kenji and Tanguy, Arnaud and Caron, Guillaume and Suzuki, Kenji and Kheddar, Abderrahmane and others},
  booktitle={2022 IEEE-RAS 21st International Conference on Humanoid Robots (Humanoids)},
  pages={306--313},
  year={2022},
  organization={IEEE}
}

@inproceedings{suomalainen2022unwinding,
  title={Unwinding rotations improves user comfort with immersive telepresence robots},
  author={Suomalainen, Markku and Sakcak, Basak and Widagdo, Adhi and Kalliokoski, Juho and Mimnaugh, Katherine J and Chambers, Alexis P and Ojala, Timo and LaValle, Steven M},
  booktitle={2022 17th ACM/IEEE International Conference on Human-Robot Interaction (HRI)},
  pages={511--520},
  year={2022},
  organization={IEEE}
}

\clearpage

\appendix
\section{Appendix}
\label{sec:appendix}
\subsection{Ablation Study}

To evaluate the contribution of each interface design component, we conduct ablation studies on the Multi-object Bimanual Retrieval and Light Switch tasks. We compare the full system against three variants: (1) \textbf{w/o Stereo}, which replaces binocular rendering with a monocular display; (2) \textbf{w/o Wide FoV}, which restricts the display to a conventional narrow egocentric field of view; and (3) \textbf{w/o Stabilization}, which removes the world-frame view stabilization and directly couples the visual feedback to robot body rotations. Tab.~\ref{tab:ablation} reports the average task completion time.

Removing \textbf{stereo vision} consistently increases completion time in both tasks. Operators have difficulties estimating the relative depth between the grippers and the target, and need multiple tries. Removing the \textbf{wide-field display} produces the largest degradation in the Multi-object Bimanual Retrieval task. With a reduced field of view, operators lose visibility of one object, requiring additional head motions and visual search. Removing \textbf{view stabilization} primarily affects the Light Switch task. Without stabilization, robot turning and locomotion continuously alter the visual reference frame, requiring additional viewpoint corrections.

\begin{table}[h]
    \centering
    \begin{tabular}{lcc}
    \toprule
    Method & Bimanual Retrieval $\downarrow$ & Light Switch $\downarrow$ 
    \\
    \hline
    Full & \textbf{9.76 $\pm$ 2.82\,s } & \textbf{11.54 $\pm$ 3.39\,s} 
    \\
    \hline
    w/o Stereo & {13.95 $\pm$ 3.60\,s } & {14.62 $\pm$ 4.43\,s } 
    \\
    \hline
    w/o Wide FoV & {14.06 $\pm$ 3.85\,s } & {17.14 $\pm$ 3.81\,s } 
    \\
    \hline
    w/o Stabilization & {10.55 $\pm$ 2.63\,s } & {18.43 $\pm$ 4.88\,s } 
    \\[-2pt]
    \bottomrule
    \end{tabular}
    \vspace{10pt}
    \caption{\textbf{Ablation Study of Interface Design Components.}}
    \label{tab:ablation}
\end{table}

Overall, the ablation study confirms that stereo perception, wide-field observation, and stabilized visual feedback provide complementary benefits for spatial perception and teleoperation efficiency.

\subsection{User Study}

We conducted a user study with 10 participants using the Multi-object Bimanual Retrieval task. Among them, one participant had prior experience with immersive egocentric VR interfaces, two participants had previous experience with exocentric robotic teleoperation systems, and the remaining seven participants had no prior experience with robot teleoperation. None of the participants were involved in this project and were unaware of the study hypothesis or which interface corresponded to the proposed system. The order of interface presentation was randomized across participants.

Participants evaluated three teleoperation interfaces: our method, Conventional Egocentric Teleoperation, and Exocentric Teleoperation. After completing the task with each interface, participants rated the system on a 5-point scale across three aspects:

\begin{itemize}
    \item \textbf{Visual Perception}: clarity of object observation, depth and distance estimation, awareness of occlusions, and the ability to maintain visual awareness of objects during the task.
    \item \textbf{Manipulation Ease}: ease of understanding spatial relationships and moving the grippers toward intended targets.
    \item \textbf{Comfort}: perceived fatigue, motion sickness, and suitability for extended teleoperation.
\end{itemize}

\begin{table}[h]
    \centering
    \begin{tabular}{lccc}
    \toprule
    Method & Visual Perception $\uparrow$ & Manipulation Ease $\uparrow$ & Comfort $\uparrow$
    \\
    \hline
    Ours & \textbf{3.4} & \textbf{3.1} & 4.0
    \\
    \hline
    Conventional Egocentric & 2.6 & 2.8 & 1.4
    \\
    \hline
    Exocentric & 1.6 & 2.2 & \textbf{4.8}
    \\[-2pt]
    \bottomrule
    \end{tabular}
    \vspace{10pt}
    \caption{\textbf{User Study Results on Visual Perception, Manipulation Ease, and Comfort.}}
    \label{tab:user_study}
\end{table}

The results are summarized in Tab.~\ref{tab:user_study}. Our method received the highest scores in both visual perception and manipulation ease. For comfort, exocentric teleoperation achieved the highest score, likely because it does not require immersive head-mounted viewing and therefore introduces less visual-motion conflict. Notably, our method was rated substantially more comfortable than conventional egocentric teleoperation, indicating that stabilized visual feedback provides preliminary subjective evidence of improved comfort while preserving the benefits of immersive first-person perception.

These findings are consistent with the quantitative results and further support the effectiveness of the proposed interface for long-horizon teleoperation and data collection.

\subsection{Experienced and Novice Operators}

To examine whether prior teleoperation experience affects the relative performance of the interfaces, we further report completion times separately for experienced and novice participants. Among the 10 participants, 3 had prior teleoperation or immersive-interface experience, while 7 were new to robot teleoperation. As shown in Tab.~\ref{tab:exp_novice}, SPOT remains competitive across both groups and achieves the lowest completion time in most settings.

\begin{table}[h]
    \centering
    \begin{tabular}{lcccc}
    \toprule
    & \multicolumn{2}{c}{Unexpected Drop} & \multicolumn{2}{c}{Peripheral Retrieval}
    \\
    & Experienced & Novice & Experienced & Novice 
    \\
    \hline
    Exo & 21.31$_{\pm 2.33}$\,s & 28.07$_{\pm 5.22}$\,s & 6.62$_{\pm 0.87}$\,s & 10.30$_{\pm 2.42}$\,s
    \\
    \hline
    Ego & 23.89$_{\pm 4.62}$\,s & 35.84$_{\pm 9.59}$\,s & 9.93$_{\pm 1.23}$\,s & 11.69$_{\pm 2.09}$\,s
    \\
    \hline
    SPOT & 15.99$_{\pm 2.32}$\,s & 21.32$_{\pm 3.53}$\,s & 6.79$_{\pm 0.77}$\,s & 7.71$_{\pm 1.64}$\,s
    \\
    \midrule
    & \multicolumn{2}{c}{Bimanual Retrieval} & \multicolumn{2}{c}{Light Switch}
    \\
    & Experienced & Novice & Experienced & Novice 
    \\
    \hline
    Exo & 9.63$_{\pm 2.07}$\,s & 15.54$_{\pm 4.03}$\,s & 16.86$_{\pm 3.83}$\,s & N/A
    \\
    \hline
    Ego & 13.40$_{\pm 2.93}$\,s & 16.34$_{\pm 3.61}$\,s & 15.71$_{\pm 2.76}$\,s & 19.20$_{\pm 4.63}$\,s
    \\
    \hline
    SPOT & 7.11$_{\pm 1.36}$\,s & 10.91$_{\pm 2.50}$\,s & 9.08$_{\pm 2.79}$\,s & 12.60$_{\pm 3.07}$\,s
    \\
    \bottomrule
    \end{tabular}
    \vspace{10pt}
    \caption{\textbf{Evaluation of experienced and novice operators.}}
    \label{tab:exp_novice}
\end{table}

\end{document}